\documentclass{article}

\usepackage[preprint]{spconf}
\usepackage{amsmath,graphicx}
\usepackage{booktabs}       
\usepackage{multirow, makecell}
\usepackage[hyphens]{url}  
\usepackage{xcolor}
\usepackage{pifont}
\usepackage{amssymb}
\usepackage{CJKutf8}
\usepackage{hyperref}
\usepackage{enumitem}

\newcommand{\cmark}{\ding{51}}%
\newcommand{\xmark}{\ding{55}}%

\newcommand{\Sref}[1]{\S\ref{#1}}

\newcommand{\Fref}[1]{Figure~\ref{#1}}

\newcommand{\Tref}[1]{Table~\ref{#1}}

\newcommand{\Hquad}{\hspace{0.3em}} 
\newcommand\mypar[1]{\noindent\textbf{#1}\Hquad}

\title{Transformer-Transducers for Code-Switched Speech Recognition}
\twoauthors
{Siddharth Dalmia\thanks{Work done while Siddharth was interning at AWS AI}}
{Carnegie Mellon University, USA \\
\texttt{sdalmia@cs.cmu.edu}}
{Yuzong Liu, Srikanth Ronanki, Katrin Kirchhoff}
{Amazon AWS AI, USA \\
\texttt{\{liuyuzon,ronanks,katrinki\}@amazon.com}}
\begin{document}
\ninept
\copyrightnotice{\copyright\ IEEE 2021}
\toappear{To appear in {\it Proc.\ ICASSP 2021, June 6-11, 2021}}

\maketitle
\begin{abstract}
We live in a world where 60\% of the population can speak two or more languages fluently. Members of these communities constantly switch between languages when having a conversation.  As automatic speech recognition (ASR) systems are being deployed to the real-world, there is a need for practical systems that can handle multiple languages both within an utterance or across utterances. 
In this paper, we present an end-to-end ASR system using a transformer-transducer model architecture for code-switched speech recognition.
We propose three modifications over the vanilla model in order to handle various aspects of code-switching. First, we introduce two auxiliary loss functions to handle the low-resource scenario of code-switching. Second, we propose a novel mask-based training strategy with language ID information to improve the label encoder training towards intra-sentential code-switching. Finally, we propose a multi-label/multi-audio encoder structure to leverage the vast monolingual speech corpora towards code-switching. We demonstrate the efficacy of our proposed approaches on the SEAME dataset, a public Mandarin-English code-switching corpus, achieving a mixed error rate of 18.5\% and 26.3\% on test$_\text{man}$ and test$_\text{sge}$ sets respectively.
\end{abstract}
\section{Introduction}
\label{sec:intro}
Code-switching (CS) refers to the phenomenon of two or more languages used by one speaker in a single conversation. CS widely exists in multilingual communities, which corresponds to around 60\% of the world's population \cite{ilanguage}. Examples include code-switching between Mandarin and English or between Spanish and English~\cite{muysken1997code, sitaram2019survey}. Code-switching can occur either at an utterance level (extra-sentential CS) or within an utterance (intra-sentential CS).

While there are numerous studies on building multilingual ASR \cite{knill2013investigation, grezl2016study, cho2018multilingual, dalmia2018sequence}, these systems typically assume that the input speech is from native speakers that do not mix different languages. However, this assumption is often impractical as speakers are bi/multilingual and continuously switch between their native language and their language of professional proficiency \cite{sitaram2019survey}. Such mixed speech poses a severe challenge to multilingual ASR systems due to different phone sets among languages, the influence of native language in pronunciation, and insufficient CS training data \cite{sitaram2019survey, flm_cm, sivasankaran-etal-2018-phone}. These effects are compounded in intra-sentential CS, which is the focus of our work.

There are promising approaches for building hybrid ASR systems for CS speech. However, these require cumbersome language-specific handcrafted features like phone merging between languages for acoustic models \cite{sivasankaran-etal-2018-phone} and linguistic structures like part-of-speech tags and language ID for language modeling \cite{flm_cm}. Additionally, the unbalanced language distribution within CS utterances can lead to a poor n-gram language model~\cite{flm_cm}, suggesting the need for handling longer contexts. End-to-end ASR systems \cite{graves2006connectionist, rnnt_graves, chan2016listen} are becoming increasingly popular, since they do not require explicit alignments and usually have fewer hyperparameters to tune. Despite their simplistic design, end-to-end ASR systems need larger amounts of training data than the hybrid based models leading to inferior performance on data-sparse tasks like code-switching \cite{Khassanov2019, zhang2020rnn}. This behavior is starting to turn around \cite{Zhou2020MultiEncoderDecoderTF} with data-augmentation techniques like SpecAugment \cite{specaugment} and joint training with alignment-based loss functions like connectionist temporal classification (CTC) loss \cite{kim2017joint}.

In this work, we propose the use of neural transducers for code-switched ASR. Unlike CTC, where each output label is conditionally independent of the others given the input speech, neural transducers condition the output on all the previous labels. Unlike attention-based encoder-decoder models, transducer models learn explicit input-output alignments, making it robust towards long utterances \cite{zeyer2020_transducer}. In particular, we focus on adapting the transformer-transducer (T-T) model to code-switched ASR \cite{zhang2020transformer_transducers, fb_tt}. The T-T model replaces the recurrent neural networks with non-recurrent multi-head self-attention transformer encoders \cite{vaswani_attention}. Transformers allow superior modeling of long-term temporal dependencies in speech data \cite{dong_speech_transformer}. As noted earlier, the ability to handle longer contexts is crucial for intra-sentential CS ASR. The language structure of a new phrase might depend on the structure before the language-switch~\cite{muysken1997code}.

\noindent We summarize the contributions of this paper below:
\begin{itemize}[leftmargin=*,itemsep=0pt, topsep=1pt]
    \item We present training strategies and insights towards improving transformer-transducer models in the data-sparse scenario of code-switching by extending the model with two auxiliary loss functions: a language model (LM) loss and a CTC loss (\Sref{sec:method_joint}). 
    \item To address intra-sentential CS, we propose  language ID (LID) aware masked training for the transformer-transducer (\Sref{sec:method_lid}). 
    \item To leverage additional monolingual corpora, we propose a multi-label/multi-audio encoder framework for the T-T model (\Sref{sec:method_monolingual}). 
\end{itemize}
On the Mandarin-English CS SEAME corpus, our proposed architecture improves over the previous RNN-transducer baseline \cite{zhang2020rnn} by around 15\% (absolute) without using any additional data and by 17\% with only 200 hours of monolingual data in each language (\Sref{sec:results}).

\section{Background and Proposed Approach}
\label{sec:methods}
\begin{figure*}[t]
  \centering
    \includegraphics[width=\linewidth]{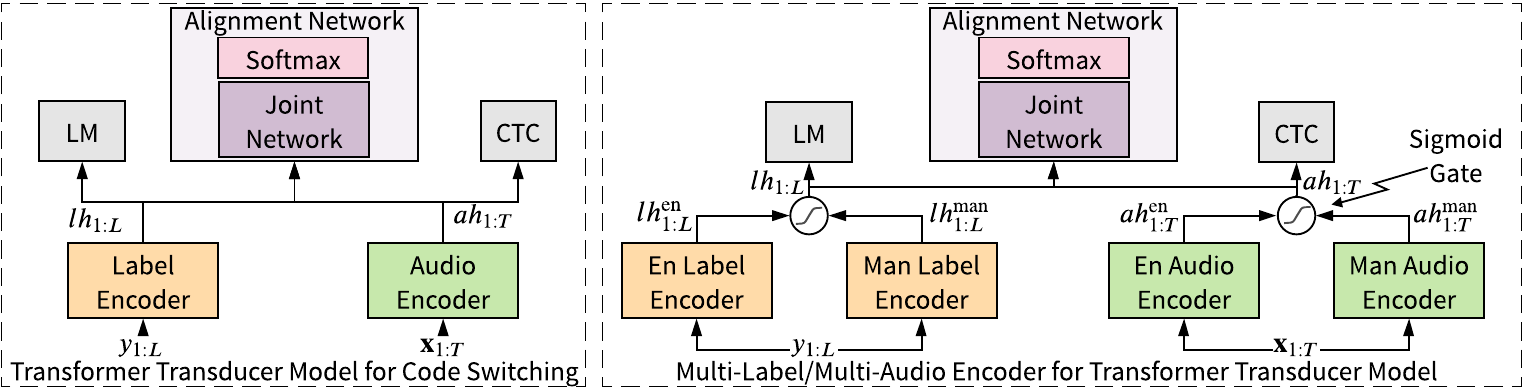}
    \caption{Proposed Transformer-Transducer Model for Code-Switching}
    \label{fig:intro}
\end{figure*}
Given an input speech sequence $\mathbf{x} = (\mathbf{x}_1,\mathbf{x}_2,\ldots,\mathbf{x}_T)$, where $\mathbf{x}_t \in \mathcal{R}^d$ is a $d$ dimensional speech feature vector and $T$ is the input sequence length, and target transcription $\mathbf{y} = (y_1, y_2,\ldots,y_L)$, where $y_l \in \mathcal{V}$ is the output label and $L$ is sequence length, the transducer loss \cite{rnnt_graves} models the posterior of the output label sequence as the marginalization over all possible alignments $\mathbf{z} \in \mathcal{Z}(\mathbf{x},\mathbf{y})$ :
\begin{align*}
P(\mathbf{y}|\mathbf{x}) &= \sum_{\mathbf{z} \in \mathcal{Z}(\mathbf{x},\mathbf{y})} P(\mathbf{z} | \mathbf{x}) 
&&= \sum_{\mathbf{z} \in \mathcal{Z}(\mathbf{x},\mathbf{y})} \prod_{i=1}^{T+L} P(z_i | \mathbf{x}, y_{1:l_{i-1}})
\end{align*}
where $\mathcal{Z}(\mathbf{x},\mathbf{y}) \subset \{\mathcal{V} \cup \phi\}^{T+L}$ corresponds to all possible values $y$ can take in the alignment path $(\mathbf{x},\mathbf{y})$ following the transducer lattice \cite{battenberg2017exploring}. A valid alignment path after removing the blank symbol $\phi$ gives the target sequence $\mathbf{y}$. $y_{1:l_{i-1}}$ corresponds to the non-blank labels chosen in the alignments till $z_i$. The transducer model consists of three components for parameterizing $P(\mathbf{z}|\mathbf{x})$ -- a label encoder, an audio encoder, and an alignment network. The label encoder encodes the output sequence $y_{1:l_{i-1}}$ as $lh$, the audio encoder encodes the input audio frames $\mathbf{x}_{1:T}$ as $ah$ and the alignment network prepares the output lattice $\mathbf{z}$ given the audio encoder and label encoder outputs using a joint network.

When transducer models were first introduced (RNN-T), the audio and label encoders used Long Short-Term Memory models (LSTMs) 
\cite{rnnt_graves}. More recently, they have been replaced with transformer encoders \cite{vaswani_attention} in the transformer-transducer model (T-T) \cite{zhang2020transformer_transducers, fb_tt}. The joint network uses feed-forward layers and a tanh non-linearity to transform the label and audio encodings to lie on orthogonal axes, resulting in an output lattice of the shape $(T, L, \mathcal{V}\cup \phi)$. For simplicity, we ignore the extra $\phi$ label encoding added to the beginning of every utterance (more details in \cite{rnnt_graves}).

Next, we describe our proposed adaptations of the transformer-transducer model to CS data. First, we present a modification to the training of the T-T model that serves to handle the `data-sparsity' and `intra-sentential' aspects of code-switching. Next, we propose multi-label and multi-audio encoders to leverage monolingual corpora. \Fref{fig:intro} presents the schematics of our proposed architecture.

\subsection{Training Transformer-Transducer for Code-Switching}
We use the vanilla T-T model as the base model towards training a CS ASR system. For all our models, we use two data augmentation strategies: three-way speed perturbation \cite{povey2011kaldi} and SpecAugment \cite{specaugment} 

\subsubsection{CTC and LM Joint Training}
\label{sec:method_joint}
Collecting code-switched data is expensive, making data-sparsity a challenge when training code-switched ASR models. 
To overcome this issue, we jointly train the audio and label encoders towards auxiliary tasks along with the standard transducer loss ($\mathcal{F}_{\mathrm{Transducer}}$) . 

The task of the audio encoder is to learn frame-level audio representations. Audio encoders trained solely with the CTC loss learns such conditionally independent frame-level representations with low amounts of training data \cite{dalmia2018sequence, tong2017multilingual}. Additionally, previous work has demonstrated the effectiveness of the CTC loss in learning better alignments when pre-training in transducers \cite{hu2020exploring_alignments} and joint-training in encoder-decoder models \cite{kim2017joint}. Taking motivation from these works, we use the CTC loss ($\mathcal{F}_{\mathrm{CTC}}$) as an auxiliary task to provide supervision to the audio encoder.

Similarly, the task of the label encoder is to encode the past context, which is used in predicting the next word during alignment with the transducer loss. Setting aside the audio alignment (which is done using the alignment network), the label encoder's task is very similar to that of the language model, allowing us to use the next word prediction task ($\mathcal{F}_{\mathrm{LM}}$), as an auxiliary task for providing supervision to the label encoder. Using the two auxiliary tasks, our overall objective function is defined as follows: 
\begin{align*}
    \mathcal{F}_{\mathrm{obj}}=\mathcal{F}_{\mathrm{Transducer} }\left(\mathbf{x}, \mathbf{y}\right)
    +\lambda_{\mathrm{CTC}} \mathcal{F}_{\mathrm{CTC}}\left(\mathbf{x}, \mathbf{y}\right) +\lambda_{\mathrm{LM}} \mathcal{F}_{\mathrm{LM}}\left(\mathbf{y}\right)
\end{align*}
where $\lambda_{\mathrm{CTC}}$ and $\lambda_{\mathrm{LM}}$ are tunable weights assigned to the auxiliary tasks. The left side of the \Fref{fig:intro}, presents the overall transducer model being used for our experiments. We use this model for all our experiments when training on code-switched data. The ablations for individual objective functions are discussed in \Sref{sec:joint_loss}.

\begin{table*}[t]
  \centering
\begin{tabular}{llcccc}
\toprule
Architecture & Model Name & Monolingual Data & dev & test$_\text{man}$ & test$_\text{sge}$ \\
\midrule
LF-MMI & Zhou et. al. (2020)~\cite{Zhou2020MultiEncoderDecoderTF} & \xmark & - & 19.0\% & 26.6\% \\
\midrule
Att Enc-Dec & Khassanov et. al. (2019)~\cite{Khassanov2019} & \xmark & - & 49.5\% & 58.9\% \\
Att Enc-Dec & Zeng et. al. (2019) ~\cite{Zeng2019} & \xmark & - & 26.4\% & 36.1\% \\
Att Enc-Dec & Zhou et. al. (2020) ~\cite{Zhou2020MultiEncoderDecoderTF} & \xmark & - & 18.9\% & 26.2\% \\
Att Enc-Dec & Our Implementation & \xmark & 20.8\% & 19.2\% & 26.9\% \\ \midrule
Transducer & Zhang et. al. (2020) ~\cite{zhang2020rnn} & \xmark & - & 33.3\% & 44.9\% \\ 
Transducer  & Our Proposed Model  & \xmark & 23.4\% & \textbf{20.2\%} & \textbf{27.7\%} \\
Transducer  & \hspace{0.5em}+Monolingual Data & \cmark & 22.2\% & \textbf{18.5\%} & \textbf{26.3\%} \\
 \bottomrule
\end{tabular}
    \caption{Results presenting the overall performance (\% MER) of our proposed transformer-transducer model. The best performing transducer models are \textbf{highlighted}. Results from previous papers and our own implementation of the Att Enc-Dec are shown for comparison.}
    \label{tab:main_results}
\end{table*}

\subsubsection{LID aware Masked Training of Label Encoder}
\label{sec:method_lid}
Intra-sentential CS can be a big challenge for transducer models as the alignment network and the label encoder have to learn alignments for different languages and learn when to switch between languages for next word prediction. To counter this, we propose two modifications to the training process of the transducer network:
\begin{itemize}[leftmargin=*,itemsep=0pt, topsep=0pt]
    \item \textbf{Randomly masking target tokens in the label encoder}: The mask tag will help the alignment network to focus on learning audio alignments independent of the target token. Additionally, in conversational speech (our target use case), the mask tags can help make the label encoder robust towards the irregularities in utterances such as disfluency and non-speech sounds. This approach could be thought of as analogous to learning back-off language models or ensembles of language models. 
    \item \textbf{Adding an LID tag to the target sequence whenever there is a switch in the language}: The LID tags help the label encoder language switches within an utterance. Additionally, they teach the alignment network which language it is currently working on.
\end{itemize}

\noindent The motivation behind the masking strategy comes from systematic dropout used for language modeling \cite{xie2017data, gal2016theoretically}. Like SpecAugment~\cite{specaugment}, we employ a similar strategy by randomly masking 40\% of the text during training. Each masked word is replaced by a \texttt{<mask>} token. For the LID tags, we add \texttt{<en>} or \texttt{<man>} tag for the start of an English segment or Mandarin segment respectively. The ablations for individual techniques will be discussed in \Sref{sec:lid_aware_training}.

\subsection{Leveraging Monolingual Corpora for Code-Switched ASR}
\label{sec:method_monolingual}
While procuring CS data is difficult, it is relatively easier to find large monolingual corpora of the languages present. However, utilizing these monolingual corpora is challenging \cite{li_ctc_2019} owing to factors like pronunciation shift, accent shift and phone influence that occur in code-switched data \cite{sivasankaran-etal-2018-phone, li_cm_am_2011}. Previous work \cite{Zhou2020MultiEncoderDecoderTF, Taneja2019} has attempted different strategies to overcome this issue. Inspired by the model in \cite{Zhou2020MultiEncoderDecoderTF}, we propose a multi-label/multi-audio encoder for handling monolingual corpora in transformer-transducer, as depicted on the right side of \Fref{fig:intro}. Here, we have individual audio and label encoders for each language present in the CS data. Each encoder accepts monolingual data in its respective language as well as CS data. The information from them are combined using a sigmoid gate $\alpha_{\text{enc}}$, with learnable weights $w_\text{enc}$ and $w_\alpha$: 
\begin{align*}
    \alpha_{\text{enc}} &= \text{sigmoid}(w_{\alpha}(\text{tanh}(w^{\text{man}}_{\text{enc}}(h^{\text{man}}) + w^{\text{en}}_{\text{enc}}(h^{\text{en}}))) \\
    h &= \alpha_{\text{enc}} * h^{\text{man}} + (1-\alpha_{\text{enc}}) * h^{\text{en}}
\end{align*}
where $h_\text{man}$ and $h_\text{en}$ are outputs of either of the label/audio encoder. We learn the gate only for code-switched data and force the gate to $1$ if the input is Mandarin only and $0$ for English only. This is done automatically for each mini-batch. This allows us to train on all the data (monolingual and code-switched) jointly end-to-end, without the need for pre-training the individual encoders, as needed in \cite{Zhou2020MultiEncoderDecoderTF}. In order to get improved results over the target dataset, the last few thousand updates are only from the CS corpora. 
We use this model when leveraging monolingual data and the contribution of the multi-label encoder and multi-audio encoder has been studied in \Sref{sec:monolingual_exp}.
     
\section{Dataset and Experimental Setup}
\label{sec:dataset}
All our transformer-transducers models are implemented using the ESPnet library \cite{espnet}. We follow the standard data prep in \cite{espnet}, where we use global mean-variance normalized $83$ log-mel filterbank and pitch features from $16$kHz audio. We augment the data with speed perturbation of $0.9$ and $1.1$. For SpecAugment, we use the SS augmentation policy presented in \cite{specaugment}. For the audio encoder, we subsample the input features by a factor of $4$ using convolutions \cite{espnet}, followed by $12$ transformer encoder blocks with $1024$ feed-forward dim and $512$ attention dim with $8$ attention heads and a dropout of $0.1$. For the label encoder, we use $4$-layer transformer encoder blocks with the same dimensions, with an attention dropout of $0.5$, a positional-embedding dropout of $0.1$, and a dropout of $0.3$ for all other components. We mask $40\%$ of the tokens in the label sequence for each utterance during training, and set $\lambda_{\text{CTC}} = 0.5$ and $\lambda_{\text{LM}} = 0.4$. We train our models with an effective mini-batch size of $192$ utterances. We use the Adam optimizer with the inverse square root decay learning rate schedule presented in \cite{espnet} with transformer-lr scale set to $2.0$ and $25$K warmup steps. We keep a validation loss patience of $5$, after which we stop training. For decoding, we use the beam-search algorithm described in \cite{rnnt_graves} with a beam size of $20$.

\mypar{SEAME Corpus:}
For our experiments, we use the SEAME corpus~\cite{Lyu2010SEAMEAM}, a conversational Mandarin-English CS corpus collected in Singapore, consisting of around 134 speakers (100 hours). We hold out 6 speakers (4.7 hours) from the training data as our development set to do hyper-parameter tuning and for studying ablations. The SEAME corpus has two official test sets, test$_\text{man}$ and test$_\text{sge}$, each consisting of 10 speakers. The test$_\text{man}$ is biased towards Mandarin speech and test$_\text{sge}$ towards English.  In the SEAME corpus, we have $\approx2.5$K Mandarin characters. We use subword-nmt \cite{subword-nmt} to convert the English target vocabulary to have a similar target size by doing $2$K byte-pair encoding (BPE) merges. We run the BPE merges after splitting English into individual words to avoid merges with Mandarin, which can be present in the context due to code-switching.

\mypar{Monolingual Corpora:} We use AISHELL-1 \cite{bu2017aishell}, a 150-hour Mandarin corpus, and  TEDLIUMv2 \cite{tedlium}, a 211-hour English corpus, as our monolingual speech dataset for the method described in \Sref{sec:method_monolingual}. We prepare $4$K English BPE units to match the $\approx4$K character set for Mandarin on the combined monolingual and CS corpora.

\mypar{Evaluation:}
We evaluate our models with Mixed Error Rate (MER), which refers to the standard WER metric but computes edits at the character-level for Mandarin and word-level for English. We use the NIST sclite scoring script to score the models and report all numbers without any post-normalization for either languages. 

\section{Results}
\label{sec:results}
\Tref{tab:main_results} presents the overall performance of our proposed transformer-transducer model. We improve over the previously published transducer model by 13.1\% and 17.2\% absolute MER over the two test sets, test$_\text{man}$ and test$_\text{sge}$, without using any monolingual training data. Our base transformer-transducer model also performs considerably better than most attention-based encoder-decoder models on this dataset and comes close to the best model \cite{Zhou2020MultiEncoderDecoderTF} and our implementation of the same. We also see that using monolingual data with the proposed multi-label encoder can improve the model further, giving us an MER of 18.5\% and 26.3\% on test$_\text{man}$ and test$_\text{sge}$ respectively.

\subsection{CTC and LM Joint Training}
\label{sec:joint_loss}
\Tref{tab:joint_loss} shows the improvements in the development set by using CTC and LM loss as auxiliary loss functions along with the transducer loss. We see that CTC loss improves the dev set MER from 25.6\% to 24.9\%. Although using LM loss causes a drop in performance, it actually helps stabilize the training of these models. With the use of LM loss, we are able to double our learning rate, reducing the convergence time from $\approx24$ hours to $\approx15$ hours. We also see in \Tref{tab:lm_joint_examples} that joint training with LM loss makes the model output grammatically coherent. This is true even though we use the LM loss during training and do not perform any LM rescoring.

\begin{table}[t]
  \centering
\begin{tabular}{lc}
\toprule
T-T Model & Dev Set \\
\midrule
Vanilla Transducer & 25.6\% \\ \midrule
\hspace{0.5em}+ CTC Loss & 24.9\% \\
\hspace{1.2em}+ LM Loss & 25.1\% \\ \midrule
\hspace{1.8em}+ MaskedTraining & 24.0\% \\
\hspace{1.8em}+ LIDMaskedTraining & 23.4\% \\
 \bottomrule
\end{tabular}
    \caption{Ablation showing the contribution of the individual proposed modifications to the vanilla transformer-transducer model.}
    \label{tab:joint_loss}
\end{table}

\begin{table}[t]
  \centering
    \begin{CJK*}{UTF8}{gkai}
\begin{tabular}{p{0.2\columnwidth} p{0.5\columnwidth}}
\toprule
\multicolumn{1}{l}{Model} & \multicolumn{1}{l}{Utterance} \\
\toprule
Reference & its like wah you waste my time \\
Base T-T  & its like wah \textcolor{red}{\textsc{we}} \textcolor{red}{\textsc{always}} my time \\
+ LM Loss & its like wah you waste my time \\ \midrule
Reference & 读 \hspace{0.05em} engineering science then 他 \\
Base T-T  & \textcolor{red}{\textsc{two}} engineering \textcolor{red}{\textsc{leh}} 他 \\
+ LM Loss & \textcolor{red}{\textsc{to}} engineering science \textcolor{red}{\textsc{and her}} \\
\bottomrule
\end{tabular}
\end{CJK*}
    \caption{Example utterances showing how jointly training with an LM auxiliary loss improves the grammar of the decoded sentence.}
    \label{tab:lm_joint_examples}
\end{table}
\begin{table}[t]
  \centering
\begin{tabular}{lc}
\toprule
T-T Model & Dev Set \\
\midrule
Proposed Transducer Model & 23.4\% \\ \midrule
\hspace{0.5em}+ Multi Audio Encoder & 23.1\% \\
\hspace{0.5em}+ Multi Label Encoder & 22.2\% \\
 \bottomrule
\end{tabular}
    \caption{Ablation showing the contribution of multi-label and multi-audio encoder when trying to leverage monolingual training data.}
    \label{tab:monolingual_exp}
\end{table}

\subsection{LID aware Masked Training}
\label{sec:lid_aware_training}
\Tref{tab:joint_loss} also presents the contribution of masked training with and without LID tags. We see that just the masked training improves the transducer model by around 1\% MER, and with the LID tags, it improves further by 0.6\% MER. We noticed that by incorporating \emph{only} LID tags, as done in \cite{zhang2020rnn}, our model has a negligible change in performance, indicating that our vanilla model is already well trained.

\subsection{Multi-Label/Multi-Audio Encoder}
\label{sec:monolingual_exp}
\Tref{tab:monolingual_exp} shows the ablation of multi-label and multi-audio encoders for leveraging monolingual training data. We see that the multi-label encoder improves considerably over our best transformer-transducer model. We do not observe significant improvement when using the multi-audio encoder likely because it is difficult to learn language boundaries in the acoustic space.
We also noticed that using both multi-audio and multi-label encoder does not cascade the improvements. We believe that using larger corpora, as in \cite{Zhou2020MultiEncoderDecoderTF}, could help overcome this issue by making the audio encoders learn better acoustic representations and expose the model to a wider set of speakers, leading to better generalizability on code-switched data.

\section{Relation to Prior Work}
\label{sec:related}
This section discusses previous literature that this work takes inspiration from and explains how our work extends from them.

\mypar{Code-Switching Background} 
The first speech recognizer for code-switched data \cite{vu_first_cs_2012} was trained on the SEAME corpus \cite{Lyu2010SEAMEAM}. They looked at phone-merging techniques to handle the two languages in acoustic modeling, explored further in \cite{sivasankaran-etal-2018-phone, li_cm_am_2011}, and generating code-switched text data for language modeling, studied more in \cite{Taneja2019,Garg2018_dlm}. Since then, different approaches have been applied to improve code-switched speech recognition like speech chains \cite{nakayama_chain_2018}, transliteration \cite{emond_transliteration_2018}, and translation \cite{huang_retraining_free_2019}. Authors in \cite{zhang2020rnn, Zeng2019,shan_investigating_2019} focus on tracking the language switch points, similar to our LID aware training. To leverage monolingual data, different techniques have been proposed like constrained output embedding \cite{Khassanov2019}, multi-encoder-decoder networks \cite{Zhou2020MultiEncoderDecoderTF} and LID integrated acoustic modeling \cite{li_ctc_2019}.

\mypar{Transducer Loss Background} 
Transducer models are widely used for online speech recognition for its streaming capabilities and low memory footprint \cite{he2019streaming, li2019improving, Huang2020ConvTransformerTL}. The transducer loss is an alignment-based loss that does the task equivalent of cross-attention in encoder-decoder models \cite{Prabhavalkar2017_comparision}, making it useful in generating closed captioning \cite{hu2020exploring_alignments, li2019improving}. With the introduction of self-attention based transformer models \cite{vaswani_attention} and data augmentation techniques like SpecAugment \cite{specaugment}, transducers have also seen competitive performance to the attention-based encoder-decoder models \cite{zhang2020transformer_transducers}.
With transducer models, CTC loss has been primarily studied as a pre-training objective \cite{zeyer2020_transducer, hu2020exploring_alignments}. This process requires a fine-tuning phase which can be cumbersome due to the decisions involved in neural-network optimization like learning rate, optimizer, etc. SpecAugment \cite{specaugment} further increases the complexity while pre-training as the audio encoder is already invariant to the noise \cite{dalmia2019enforcing}. The label encoder in transducers behaves like a language model that can benefit from large amounts of text data \cite{rnnt_graves}. Due to the code-switching nature of our task, obtaining text data is difficult; instead, we use the next word prediction task as a joint training objective.

\section{Conclusion}
\label{sec:conclusion}
In this paper, we present a transformer-transducer model for code-switched speech recognition. We show significant improvements over the previous transducer model and perform at par with the best attention-based encoder-decoder and LF-MMI based hybrid models. We propose modifications to improve transformer-transducers training towards the data-sparse and intra-sentential nature of code-switched corpora. Additionally, we propose a multi-label/audio encoder framework to leverage monolingual data to improve recognition performance. In the future, we would like to extend this mechanism to further train with unlabeled audio from either monolingual or code-switched sources in an unsupervised manner.

\section{Acknowledgements}
\label{sec:ack}
We are grateful to the AWS Speech Science team, Shruti Rijhwani, Samridhi Choudhary and Brian Yan for their valuable feedback.

\footnotesize
\bibliographystyle{IEEEbib}
\bibliography{refs}

\begin{thebibliography}{10}

\bibitem{ilanguage}
``{Multilingual People},'' \url{http://ilanguages.org/bilingual.php},
\newblock Accessed: 2020-10-15.

\bibitem{muysken1997code}
P.~Muysken,
\newblock ``{Code-switching processes: Alternation, insertion, congruent
  lexicalization},''
\newblock {\em Language choices: Conditions, constraints, and consequences},
  1997.

\bibitem{sitaram2019survey}
S.~Sitaram, K.~R. Chandu, S.~K. Rallabandi, and A.~W Black,
\newblock ``{A Survey of Code-switched Speech and Language Processing},''
\newblock {\em arXiv:1904.00784}, 2019.

\bibitem{knill2013investigation}
K.~Knill, M.~Gales, S.~P Rath, P.~C Woodland, C.~Zhang, and S.~Zhang,
\newblock ``{Investigation of multilingual deep neural networks for spoken term
  detection},''
\newblock in {\em Proc. ASRU}, 2013.

\bibitem{grezl2016study}
F.~Gr{\'e}zl, E.~Egorova, and M.~Karafi{\'a}t,
\newblock ``{Study of large data resources for multilingual training and system
  porting},''
\newblock in {\em Proc. SLTU}, 2016.

\bibitem{cho2018multilingual}
J.~Cho, M.~K. Baskar, R.~Li, M.~Wiesner, S.~H. Mallidi, et~al.,
\newblock ``{Multilingual sequence-to-sequence speech recognition:
  architecture, transfer learning, and language modeling},''
\newblock in {\em Proc. SLT}, 2018.

\bibitem{dalmia2018sequence}
S.~Dalmia, R.~Sanabria, F.~Metze, and A.~W. Black,
\newblock ``{Sequence-based Multi-lingual Low Resource Speech Recognition},''
\newblock in {\em Proc. ICASSP}, 2018.

\bibitem{flm_cm}
H.~Adel, N.~T. Vu, and T.~Schultz,
\newblock ``{Combination of Recurrent Neural Networks and Factored Language
  Models for Code-Switching Language Modeling},''
\newblock in {\em Proc. ACL}, 2013.

\bibitem{sivasankaran-etal-2018-phone}
S.~Sivasankaran, B.~M.~L. Srivastava, S.~Sitaram, et~al.,
\newblock ``{Phone Merging For Code-Switched Speech Recognition},''
\newblock in {\em Workshop on Computational Approaches to Linguistic
  Code-Switching}, 2018.

\bibitem{graves2006connectionist}
A.~Graves, S.~Fern{\'a}ndez, F.~Gomez, and J.~Schmidhuber,
\newblock ``{Connectionist Temporal Classification: Labelling Unsegmented
  Sequence Data with Recurrent Neural Networks},''
\newblock in {\em Proc. ICML}, 2006.

\bibitem{rnnt_graves}
A.~Graves,
\newblock ``{Sequence Transduction with Recurrent Neural Networks},''
\newblock in {\em ICML Representation Learning Workshop}, 2012.

\bibitem{chan2016listen}
W.~Chan, N.~Jaitly, Q.~Le, and O.~Vinyals,
\newblock ``{Listen, attend and spell: A neural network for large vocabulary
  conversational speech recognition},''
\newblock in {\em Proc. ICASSP}, 2016.

\bibitem{Khassanov2019}
Y.~Khassanov, H.~Xu, V.~T. Pham, Z.~Zeng, E.~S. Chng, et~al.,
\newblock ``{Constrained Output Embeddings for End-to-End Code-Switching Speech
  Recognition with Only Monolingual Data},''
\newblock in {\em Proc. Interspeech}, 2019.

\bibitem{zhang2020rnn}
S.~Zhang, J.~Yi, Z.~Tian, J.~Tao, and Y.~Bai,
\newblock ``{RNN-Transducer with language bias for end-to-end Mandarin-English
  code-switching speech recognition},''
\newblock {\em arXiv:2002.08126}, 2020.

\bibitem{Zhou2020MultiEncoderDecoderTF}
X.~Zhou, E.~Yilmaz, Y.~Long, Y.~Li, and H.~Li,
\newblock ``{Multi-Encoder-Decoder Transformer for Code-Switching Speech
  Recognition},''
\newblock in {\em Proc. Interspeech}, 2020.

\bibitem{specaugment}
D.~S. Park, W.~Chan, Y.~Zhang, C.~Chiu, et~al.,
\newblock ``{SpecAugment: A Simple Data Augmentation Method for Automatic
  Speech Recognition},''
\newblock in {\em Proc. Interspeech}, 2019.

\bibitem{kim2017joint}
S.~Kim, T.~Hori, and S.~Watanabe,
\newblock ``{Joint CTC-attention based end-to-end speech recognition using
  multi-task learning},''
\newblock in {\em Proc. ICASSP}, 2017.

\bibitem{zeyer2020_transducer}
A.~Zeyer, A.~Merboldt, R.~Schl{\"u}ter, and H.~Ney,
\newblock ``{A New Training Pipeline for an Improved Neural Transducer},''
\newblock {\em arXiv:2005.09319}, 2020.

\bibitem{zhang2020transformer_transducers}
Q.~Zhang, H.~Lu, H.~Sak, A.~Tripathi, et~al.,
\newblock ``{Transformer transducer: A streamable speech recognition model with
  transformer encoders and RNN-T loss},''
\newblock in {\em Proc. ICASSP}, 2020.

\bibitem{fb_tt}
C.~Yeh, J.~Mahadeokar, K.~Kalgaonkar, Y.~Wang, D.~Le, et~al.,
\newblock ``{Transformer-Transducer: End-to-End Speech Recognition with
  Self-Attention},''
\newblock {\em arXiv:1910.12977}, 2019.

\bibitem{vaswani_attention}
A.~Vaswani, N.~Shazeer, N.~Parmar, J.~Uszkoreit, L.~Jones, et~al.,
\newblock ``{Attention is All you Need},''
\newblock in {\em Proc. NeurIPS}, 2017.

\bibitem{dong_speech_transformer}
L.~{Dong}, S.~{Xu}, and B.~{Xu},
\newblock ``{Speech-Transformer: A No-Recurrence Sequence-to-Sequence Model for
  Speech Recognition},''
\newblock in {\em Proc. ICASSP}, 2018.

\bibitem{battenberg2017exploring}
E.~Battenberg, J.~Chen, R.~Child, A.~Coates, et~al.,
\newblock ``{Exploring neural transducers for end-to-end speech recognition},''
\newblock in {\em Proc. ASRU}, 2017.

\bibitem{povey2011kaldi}
D.~Povey, A.~Ghoshal, G.~Boulianne, L.~Burget, et~al.,
\newblock ``{The Kaldi Speech Recognition Toolkit},''
\newblock in {\em Proc. ASRU}, 2011.

\bibitem{tong2017multilingual}
S.~Tong, P.~N. Garner, and H.~Bourlard,
\newblock ``{Multilingual training and cross-lingual adaptation on CTC-based
  acoustic model},''
\newblock {\em arXiv:1711.10025}, 2017.

\bibitem{hu2020exploring_alignments}
H.~Hu, R.~Zhao, J.~Li, L.~Lu, and Y.~Gong,
\newblock ``{Exploring Pre-Training with Alignments for RNN Transducer Based
  End-to-End Speech Recognition},''
\newblock in {\em Proc. ICASSP}, 2020.

\bibitem{Zeng2019}
Z.~Zeng, Y.~Khassanov, V.~T. Pham, H.~Xu, E.~S. Chng, and H.~Li,
\newblock ``{On the End-to-End Solution to Mandarin-English Code-Switching
  Speech Recognition},''
\newblock in {\em Proc. Interspeech}, 2019.

\bibitem{xie2017data}
Z.~Xie, S.~I Wang, J.~Li, D.~L{\'e}vy, A.~Nie, et~al.,
\newblock ``{Data Noising as Smoothing in Neural Network Language Models},''
\newblock in {\em ICLR}, 2017.

\bibitem{gal2016theoretically}
Y.~Gal and Z.~Ghahramani,
\newblock ``{A Theoretically Grounded Application of Dropout in Recurrent
  Neural Networks},''
\newblock in {\em Proc. NeurIPS}, 2016.

\bibitem{li_ctc_2019}
K.~{Li}, J.~{Li}, G.~{Ye}, R.~{Zhao}, and Y.~{Gong},
\newblock ``{Towards Code-switching ASR for End-to-end CTC Models},''
\newblock in {\em Proc. ICASSP}, 2019.

\bibitem{li_cm_am_2011}
Y.~{Li}, P.~{Fung}, P.~{Xu}, and Y.~{Liu},
\newblock ``{Asymmetric acoustic modeling of mixed language speech},''
\newblock in {\em Proc. ICASSP}, 2011.

\bibitem{Taneja2019}
K.~Taneja, S.~Guha, P.~Jyothi, and B.~Abraham,
\newblock ``{Exploiting Monolingual Speech Corpora for Code-Mixed Speech
  Recognition},''
\newblock in {\em Proc. Interspeech}, 2019.

\bibitem{espnet}
S.~Watanabe, T.~Hori, S.~Karita, T.~Hayashi, et~al.,
\newblock ``{ESPnet: End-to-End Speech Processing Toolkit},''
\newblock in {\em Proc. Interspeech}, 2018.

\bibitem{Lyu2010SEAMEAM}
D.~Lyu, T.~Tan, C.~Siong, and H.~Li,
\newblock ``{SEAME: a Mandarin-English code-switching speech corpus in
  south-east asia},''
\newblock in {\em Proc. Interspeech}, 2010.

\bibitem{subword-nmt}
R.~Sennrich, B.~Haddow, and A.~Birch,
\newblock ``{Neural Machine Translation of Rare Words with Subword Units},''
\newblock in {\em Proc. ACL}, 2016.

\bibitem{bu2017aishell}
H.~Bu, J.~Du, X.~Na, B.~Wu, and H.~Zheng,
\newblock ``{AISHELL-1: An Open-Source Mandarin Speech Corpus and A Speech
  Recognition Baseline},''
\newblock in {\em Proc. O-COCOSDA}, 2017.

\bibitem{tedlium}
A.~Rousseau, P.~Del{\'e}glise, and Y.~Est{\`e}ve,
\newblock ``{Enhancing the TED-LIUM Corpus with Selected Data for Language
  Modeling and More TED Talks},''
\newblock in {\em Proc. LREC}, 2014.

\bibitem{vu_first_cs_2012}
N.~T. {Vu}, D.~{Lyu}, J.~{Weiner}, et~al.,
\newblock ``{A first speech recognition system for Mandarin-English code-switch
  conversational speech},''
\newblock in {\em Proc. ICASSP}, 2012.

\bibitem{Garg2018_dlm}
S.~Garg, T.~Parekh, and P.~Jyothi,
\newblock ``{Dual Language Models for Code Switched Speech Recognition},''
\newblock in {\em Proc. Interspeech}, 2018.

\bibitem{nakayama_chain_2018}
S.~{Nakayama}, A.~{Tjandra}, S.~{Sakti}, and S.~{Nakamura},
\newblock ``{Speech Chain for Semi-Supervised Learning of Japanese-English
  Code-Switching ASR and TTS},''
\newblock in {\em Proc. SLT}, 2018.

\bibitem{emond_transliteration_2018}
J.~{Emond}, B.~{Ramabhadran}, B.~{Roark}, et~al.,
\newblock ``{Transliteration Based Approaches to Improve Code-Switched Speech
  Recognition Performance},''
\newblock in {\em Proc. SLT}, 2018.

\bibitem{huang_retraining_free_2019}
Z.~{Huang}, X.~{Zhuang}, D.~{Liu}, et~al.,
\newblock ``{Exploring Retraining-free Speech Recognition for Intra-sentential
  Code-switching},''
\newblock in {\em Proc. ICASSP}, 2019.

\bibitem{shan_investigating_2019}
C.~{Shan}, C.~{Weng}, G.~{Wang}, et~al.,
\newblock ``{Investigating End-to-end Speech Recognition for Mandarin-english
  Code-switching},''
\newblock in {\em Proc. ICASSP}, 2019.

\bibitem{he2019streaming}
Y.~He, T.~N. Sainath, R.~Prabhavalkar, et~al.,
\newblock ``{Streaming End-to-end Speech Recognition For Mobile Devices},''
\newblock in {\em Proc. ICASSP}, 2019.

\bibitem{li2019improving}
J.~Li, R.~Zhao, H.~Hu, and Y.~Gong,
\newblock ``{Improving RNN Transducer Modeling for End-to-End Speech
  Recognition},''
\newblock in {\em Proc. ASRU}, 2019.

\bibitem{Huang2020ConvTransformerTL}
W.~Huang, W.~Hu, Y.~Yeung, and X.~Chen,
\newblock ``{Conv-Transformer Transducer: Low Latency, Low Frame Rate,
  Streamable End-to-End Speech Recognition},''
\newblock in {\em Proc. Interspeech}, 2020.

\bibitem{Prabhavalkar2017_comparision}
R.~Prabhavalkar, K.~Rao, T.~N. Sainath, et~al.,
\newblock ``{A Comparison of Sequence-to-Sequence Models for Speech
  Recognition},''
\newblock in {\em Proc. Interspeech}, 2017.

\bibitem{dalmia2019enforcing}
S.~Dalmia, A.~Mohamed, M.~Lewis, et~al.,
\newblock ``{Enforcing encoder-decoder modularity in sequence-to-sequence
  models},''
\newblock {\em arXiv:1911.03782}, 2019.

\end{thebibliography}

\end{document}